\def\year{2017}\relax
\documentclass[letterpaper]{article}
\usepackage{arxiv}
\usepackage{times}
\usepackage{helvet}
\usepackage{courier}
\usepackage{url}
\usepackage{amsmath}
\usepackage{amssymb}
\usepackage{booktabs}
\usepackage{graphicx}
\usepackage{listings}
\usepackage{paralist}
\usepackage{subfig}
\usepackage{epstopdf}
\usepackage{epsfig}
\usepackage{underscore}

\title{Semantic Parsing of Mathematics by\\ Context-based Learning from Aligned Corpora and Theorem Proving}
\author{Cezary Kaliszyk\thanks{Supported by the Austrian Science Fund (FWF): P26201.} \\ University of Innsbruck, Austria
\AND 
Josef Urban\thanks{Supported by the ERC Consolidator grant no.\ 649043
\textit{AI4REASON}.} \and Ji\v{r}\'i Vysko\v{c}il\footnotemark[2] \\ Czech Technical University in Prague, Czech Republic
}

\begin{document}

\maketitle
%
\begin{abstract}
  We study methods for automated parsing of informal mathematical
  expressions into formal ones, a main prerequisite for deep computer
  understanding of informal mathematical texts. We propose a
  context-based parsing approach that combines efficient statistical
  learning of deep parse trees with their semantic pruning by type
  checking and large-theory automated theorem proving. We show that
  the methods very significantly improve on previous results in
  parsing theorems from the Flyspeck corpus.
\end{abstract}

\section{Introduction}
\label{intro}
Computer-understandable (formal) mathematics~\cite{HarrisonUW14} is still far
from taking over the mathematical mainstream. Despite recent impressive
formalizations such as the Formal Proof of the Kepler conjecture (Flyspeck)~\cite{HalesABDHHKMMNNNOPRSTTTUVZ15}, Feit-Thompson~\cite{DBLP:conf/itp/GonthierAABCGRMOBPRSTT13}, seL4~\cite{KleinAEHCDEEKNSTW10}, CompCert~\cite{Leroy09}, and
CCL~\cite{BancerekR02}, formalizing proofs is still largely unappealing to
mathematicians. While research on AI and 
strong automation over
large theories has taken off in the last decade \cite{hammers4qed},
there has been
so far 
little progress in automating the understanding of
informal \LaTeX-written and ambiguous mathematical writings.

Automatic parsing of informal mathematical texts into formal ones has
been for long time considered a hard or impossible task. Among 
the state-of-the-art Interactive Theorem Proving (ITP) systems such as HOL
(Light)~\cite{Harrison96}, Isabelle~\cite{WenzelPN08},
Mizar~\cite{mizar-in-a-nutshell} and Coq~\cite{coq}, none includes
automated parsing, instead relying on sophisticated formal 
languages and mechanisms~\cite{GarillotGMR09,GonthierT12,HaftmannW06,RudnickiST01}. The
past work in this direction -- most notably by Zinn~\cite{Zinn2004} -- has often 
been cited as discouraging from such efforts.

Recently~\cite{KaliszykUV15} proposed to automatically learn
formal understanding of informal mathematics from large aligned
informal/formal corpora. Such learning can be
additionally combined with strong semantic filtering methods such as
typechecking and large-theory Automated Theorem Proving (ATP).
Suitable aligned corpora are starting to
appear today, the major example being the Flyspeck project and in particular
its alignment (by Hales) with the detailed informal Blueprint for
Formal Proofs~\cite{hales-dense}.


\subsection{Contributions}
In this paper, we first introduce the informal-to-formal setting (Sec.~\ref{setting}), summarize the probabilistic context-free grammar (PCFG) approach of \cite{KaliszykUV15} (Sec.~\ref{sec:align}),
and extend this approach by fast context-aware parsing mechanisms.
\begin{itemize}
\item \textbf{Limits of the context-free approach.} We demonstrate on a minimal example, that the context-free
  setting is not strong enough to eventually learn correct parsing
  (Sec.~\ref{sec:example}) of relatively simple informal mathematical formulas.
\item \textbf{Efficient context inclusion via discrimination trees.} 
We propose and efficiently implement modifications of
  the CYK algorithm that take into account larger parsing subtrees
  (context) and their probabilities
  (Sec.~\ref{sec:subtrees}). 
This modification is motivated
  by an analogy with large-theory reasoning systems and its efficient
  implementation is based on a novel use of fast theorem-proving data structures that extend the probabilistic parser.
\item \textbf{Significant improvement of the informal-to-formal translation performance}. 
The methods are evaluated, both by standard (non-semantic)
  machine-learning cross-validation, and by strong semantic methods
  available in formal mathematics such as typechecking and
  large-theory automated reasoning (Sec.~\ref{sec:evaluation}).

\end{itemize}


\section{Informalized Flyspeck and PCFG}
\label{setting}
The ultimate goal of the informal-to-formal traslation is to
automatically learn parsing on informal \LaTeX{} formulas that have
been aligned with their formal counterparts, as for example done by
Hales for his informal and formal Flyspeck
texts~\cite{hales-dense,FlyspeckWiki}.  Instead of starting with
\LaTeX{} where only hundreds of aligned examples are so far available for Flyspeck, we re-use the first large informal/formal corpus introduced
in~\cite{KaliszykUV15}, based on \textit{informalized} (or
\textit{ambiguated}) formal statements created from the HOL Light theorems in Flyspeck. 
This provides about 22000 informal/formal pairs of Flyspeck theorems.
\subsection{Informalized Flyspeck}
The following transformations are applied in~\cite{KaliszykUV15} to the HOL parse trees to obtain the aligned corpus:
\begin{itemize}
\item Using the 72 overloaded instances defined in HOL Light/Flyspeck, such as
 \texttt{("+", "vector_add")}.
The constant \texttt{vector_add} would be replaced by \texttt{+} in the resulting sentence.
\item Getting the infix operators from HOL Light, and printing them as infix
in the informalized sentences. Since \texttt{+} is declared as infix, \texttt{vector_add u v}, would thus result in \texttt{u + v}.
\item Getting all ``prefixed'' symbols from the list of 1000 most frequent symbols by searching for:
\texttt{real_, int_, vector_, nadd_, treal_, hreal_, matrix_, complex_}
and making them ambiguous by forgetting the prefix.
\item Similar overloading of various other symbols that disambiguate overloading, for example the ``c''-versions of functions such as \texttt{ccos cexp clog csin}, 
similarly for
 \texttt{vsum, rpow, nsum, list_sum}, etc. 
\item Deleting brackets, type annotations, and the 10 most frequent casting functors such as \texttt{Cx} and \texttt{real_of_num}.
\end{itemize}

\subsection{The Informal-To-Formal Translation Task}
The \textit{informal-to-formal translation task} is to construct an AI system that will automatically 
produce the most probable formal (in this case HOL) parse trees for previously unseen informal sentences. 
 For example, the informalized statement of the HOL theorem \texttt{REAL_NEGNEG}:
\begin{quote}
\begin{small}
\texttt{
! A0 -\-- -\-- A0 = A0} 
\end{small}
\end{quote}
has the formal HOL Light representation shown in
Fig. \ref{HOL11} (as a text) and in Fig. \ref{fig:origtree1} (as a tree).  Note that
all overloaded symbols are disambiguated there, they are applied with
the correct arity, and all terms are decorated with their result
types. To solve the task, we allow (and assume) training on a
sufficiently large corpus of such informal/formal pairs.

\begin{figure}[bth]
\begin{scriptsize}
\texttt{(Comb (Const "!" (Tyapp "fun" (Tyapp "fun" (Tyapp "real") (Tyapp "bool")) (Tyapp "bool"))) (Abs "A0" (Tyapp "real") (Comb (Comb (Const "=" (Tyapp "fun" (Tyapp "real") (Tyapp "fun" (Tyapp "real") (Tyapp "bool")))) (Comb (Const "real_neg" 
(Tyapp "fun" (Tyapp "real") (Tyapp "real"))) (Comb (Const "real_neg" (Tyapp "fun" (Tyapp "real") (Tyapp "real"))) (Var "A0" (Tyapp "real"))))) 
(Var "A0" (Tyapp "real")))))}
\end{scriptsize}
\caption{\label{HOL11}The HOL Light representation of \texttt{REAL_NEGNEG}}
\end{figure}

\subsection{Probabilistic Context Free Grammars}
\label{sec:pcfg}
Given a large corpus of corresponding informal/formal
formulas, how can we
train an AI system for parsing the next informal formula into a formal one?
The informal-to-formal domain differs from natural-language
domains, where millions of examples of paired (e.g., English/German)
sentences are available for training machine translation. The natural
languages also have many more words (concepts) than in mathematics,
and the sentences to a large extent also lack the recursive structure
that is frequently encountered in mathematics. Given that 
there are currently only thousands of informal/formal examples, purely
statistical alignment methods based on n-grams seem
inadequate. Instead, the methods have to learn how to compose larger
parse trees from smaller ones based on those encountered in the
limited number of examples.

A well-known approach ensuring such compositionality is the
use of CFG (Context Free Grammar) parsers. This approach has been
widely used, e.g., in word-sense disambiguation. 
A frequently used CFG algorithm is the CYK (Cocke--Younger--Kasami) chart-parser~\cite{Younger67},
based on bottom-up parsing. 
By default CYK
requires the CFG to be in the Chomsky Normal Form (CNF). The
transformation to CNF can cause an exponential blow-up of the
grammar, however, an improved version of CYK gets
around this issue~\cite{DBLP:journals/didactica/LangeL09}.
 
In linguistic applications the input grammar for the CFG-based parsers is typically extracted
from the \emph{grammar trees} which correspond to the correct parses of
natural-language sentences.  Large annonated \emph{treebanks} of such correct
parses exist for natural languages. The
grammar rules extracted from the treebanks are typically ambiguous:
there are multiple possible parse trees for a particular
sentence. This is why CFG is extended by adding a probability to each
grammar rule, resulting in Probabilistic CFG (PCFG). 



\section{PCFG for the Informal-To-Formal Task}
\label{sec:align}

The most straightforward PCFG-based approach would be to
directly use the native HOL Light parse trees (Fig. \ref{fig:origtree1}) for extracting the PCFG.
However, terms and types are there annotated with only a few nonterminals such as:
\texttt{Comb} (application), \texttt{Abs} (abstraction),
\texttt{Const} (higher-order constant), \texttt{Var} (variable),
\texttt{Tyapp} (type application), and \texttt{Tyvar} (type variable).
This would lead to many possible parses in the context-free setting, because the learned rules are very universal, e.g:
\begin{quote}
\begin{footnotesize}
\texttt{Comb -> Const Var.}  \texttt{Comb -> Const Const.} \texttt{Comb -> Comb Comb.}
\end{footnotesize}
\end{quote}
The type information does not help to constrain the applications, and
the last rule allows a series of several constants to be given
arbitrary application order, leading to uncontrolled explosion.

\subsection{HOL Types as Nonterminals}
\begin{figure}[bth]
 \centering
 \includegraphics[width=9cm]{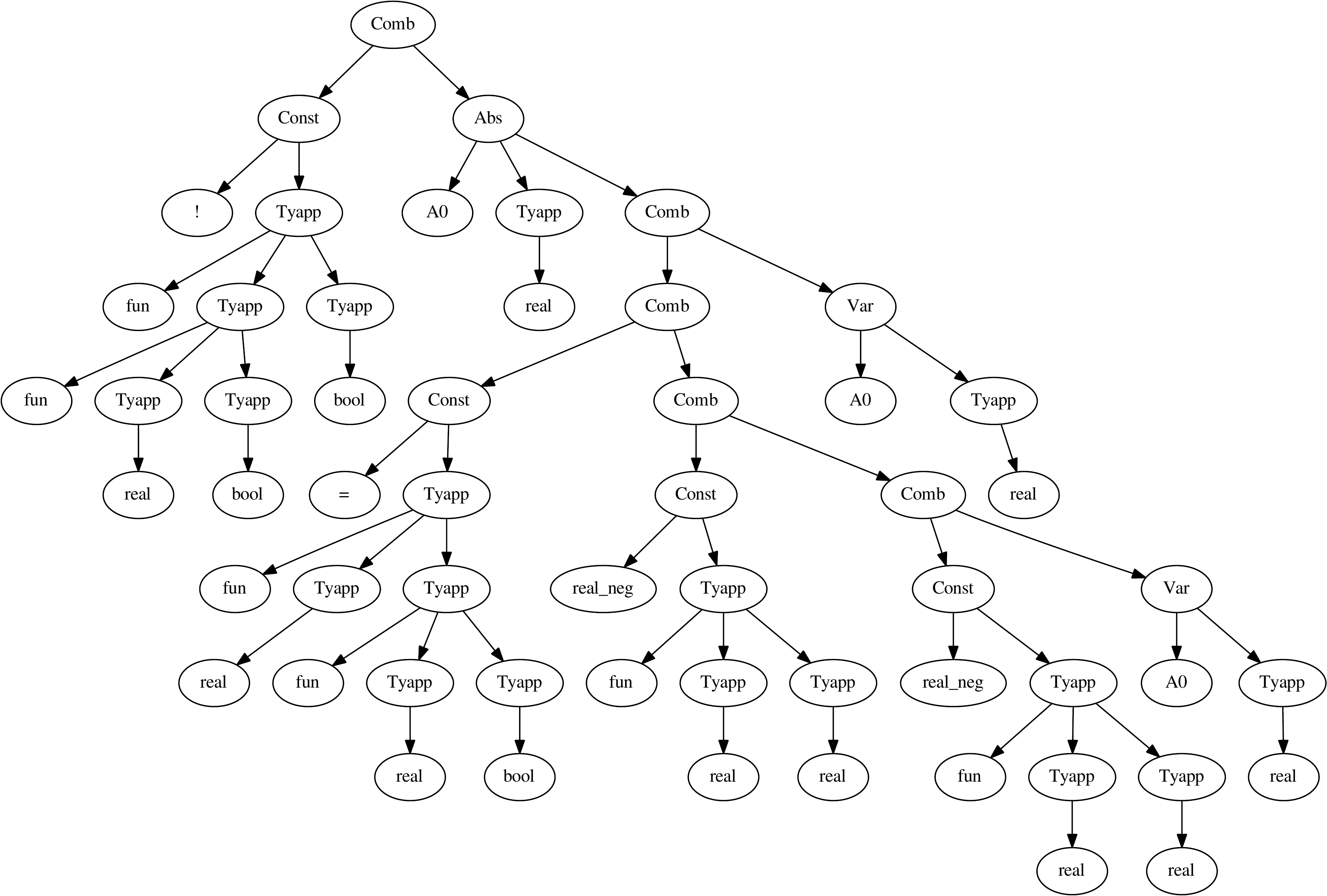}
 \caption{The HOL Light parse tree of \texttt{REAL_NEGNEG}}
 \label{fig:origtree1}
 \end{figure}

The approach taken in~\cite{KaliszykUV15} is to first
re-order and simplify the HOL Light
parse trees to propagate the type information at appropriate places.
This gives the context-free rules a chance of providing meaningful pruning information. 
For example, consider again the raw HOL Light parse tree for   \texttt{REAL_NEGNEG} (Fig.~\ref{HOL11},\ref{fig:origtree1}).

\begin{figure}[!htb]
 \centering
 \includegraphics[height=6cm,scale=.5]{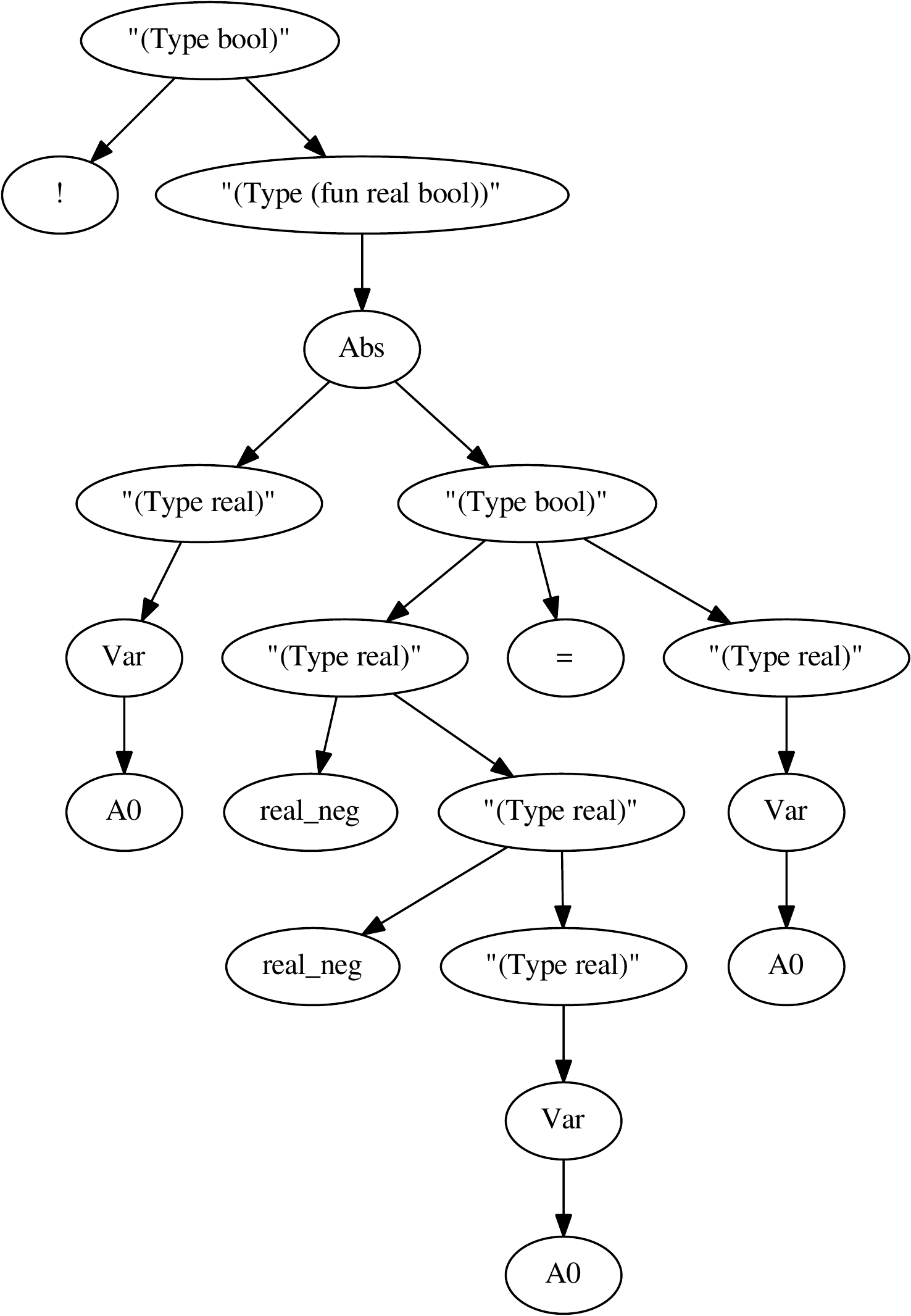}
 \caption{Transformed tree of \texttt{REAL_NEGNEG}}
 \label{fig:origtree2}
 \end{figure}
 Instead of directly extracting very general rules such as \texttt{Comb -> Const Abs}, each type is first
 compressed into an opaque nonterminal. This turns the parse tree of
 \texttt{REAL_NEGNEG} into (see also Fig.~\ref{fig:origtree2}):
\begin{quote}
\begin{scriptsize}
\texttt{("(Type bool)" ! ("(Type (fun real bool))" (Abs ("(Type real)" (Var A0)) ("(Type bool)" ("(Type real)" real_neg ("(Type real)" real_neg ("(Type real)" (Var A0)))) = ("(Type real)" (Var A0))))))}
\end{scriptsize}
\end{quote}
The CFG rules extracted from this transformed tree thus become more targeted. For example, the two rules:
\begin{quote}
\begin{scriptsize}
\texttt{"(Type bool)" -> "(Type real)" = "(Type real)".}\\
\texttt{"(Type real)" -> real_neg "(Type real)".}
\end{scriptsize}
\end{quote}
say that equality of two reals has type \texttt{bool}, and negation
applied to reals yields reals. 
Such learned \emph{probabilistic typing rules} restrict the
number of possible parses much more than the general ``application''
rules extracted from the original HOL Light tree.
The rules still have a
non-trivial generalization (learning) effect that is needed for the
compositional behavior of the information extracted from the
trees. For example, once we learn from the training data that the variable \texttt{``u''} is mostly
parsed as a real number, i.e.:
\begin{quote}
\begin{scriptsize}
\texttt{"(Type real)" -> Var u.}
\end{scriptsize}
\end{quote}
we will be able to apply \texttt{real_neg} to \texttt{``u''} even if the
particular subterm \texttt{``real_neg u''} has never yet been seen in
the training examples, and the probability of this parse will be
relatively high. 

In other words, having the HOL types as
\emph{semantic categories} (corresponding e.g. to word senses when
using PCFG for word-sense disambiguation) is a reasonable choice for
the first experiments. It is however likely that even better semantic
categories can be developed, based on more involved statistical and
semantic analysis of the data such as \emph{latent
  semantics}~\cite{DeerwesterDLFH90}.

\subsection{Semantic Concepts as Nonterminals}
\begin{figure}[!htb]
 \centering
 \includegraphics[scale=.24]{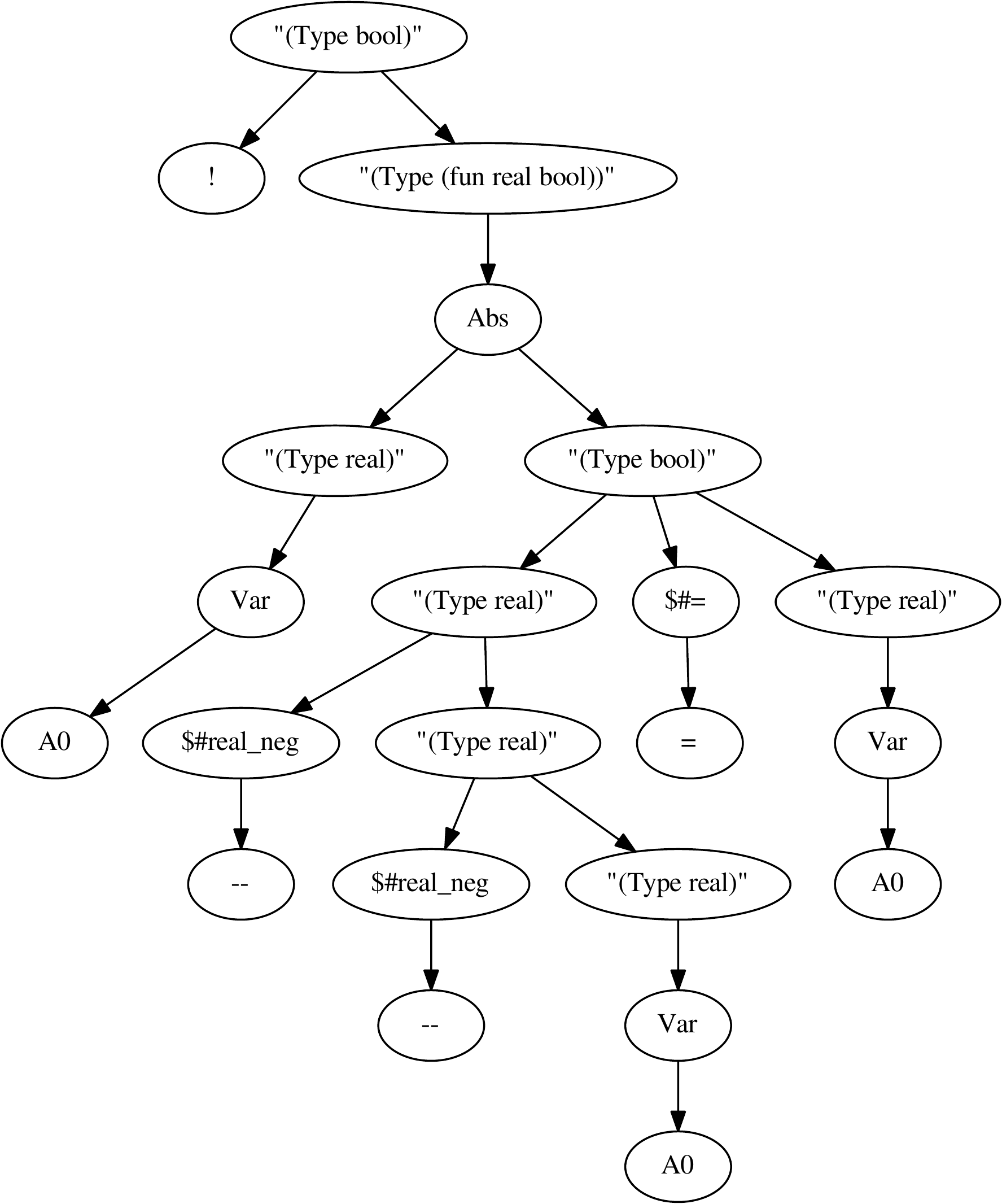}
 \caption{The tree of \texttt{REAL_NEGNEG} used for actual grammar training}
 \label{fig:newtree}
 \end{figure}

The last part of the original setting wraps ambiguous symbols, such as \texttt{``--''}, 
in their disambiguated \emph{semantic/formal concept}
nonterminals. In this case \texttt{\$\#real_neg} would be wrapped around  \texttt{``--''} in the training tree when  \texttt{``--''} is used as subtraction on reals. 
While the type
annotation is often sufficient for disambiguation, such explicit
disambiguation nonterminal is more precise and allows
easier extraction of the HOL semantics from the constructed parse
trees. The actual tree of \texttt{REAL_NEGNEG} used for training the grammar is thus as follows (see also Fig.~\ref{fig:newtree}):
\begin{quote}
\begin{scriptsize}
\texttt{("(Type bool)" ! ("(Type (fun real bool))" (Abs ("(Type real)" (Var A0)) ("(Type bool)" ("(Type real)" (\$\#real_neg --) ("(Type real)" (\$\#real_neg --) ("(Type real)" (Var A0)))) (\$\#= =) ("(Type real)" (Var A0))))))}
\end{scriptsize}
\end{quote}

\subsection{Modified CYK Parsing and Its Initial Performance}
 Once the PCFG is learned from such data, the CYK algorithm augmented
 with fast internal semantic checks is used to parse the
 informal sentences. The semantic checks are performed to require
 compatibility of the types of free variables in parsed subtrees. The
 most probable parse trees are then typechecked by HOL Light. This is followed by proof and disproof attempts by the HOL(y)Hammer
 system~\cite{holyhammer}, using all the semantic knowledge available in the Flyspeck
 library (about 22000 theorems). 
The first
 large-scale disambiguation experiment conducted over ``ambiguated''
 Flyspeck in~\cite{KaliszykUV15} showed that about 40\% of the ambiguous
 sentences have their correct parses among the best 20 parse trees
 produced by the trained parser. This is encouraging, but certainly
 invites further research in improving the statistical/semantic parsing methods.

\section{Limits of the Context-Free Grammars}\label{sec:example}
A major limiting issue when using PCFG-based parsing algorithms is the
context-freeness of the grammar. This is most obvious when using just the low-level term constructors as nonterminals,
however it shows often also in the more advanced setting described above.
In some cases, no matter how good are
the training data, there is no way how to set up the probabilities of the parsing rules 
 so that the required parse tree will have the highest
probability. We show this on the following simple example.

\begin{figure}[!htb]
 \centering
 \includegraphics[scale=.25]{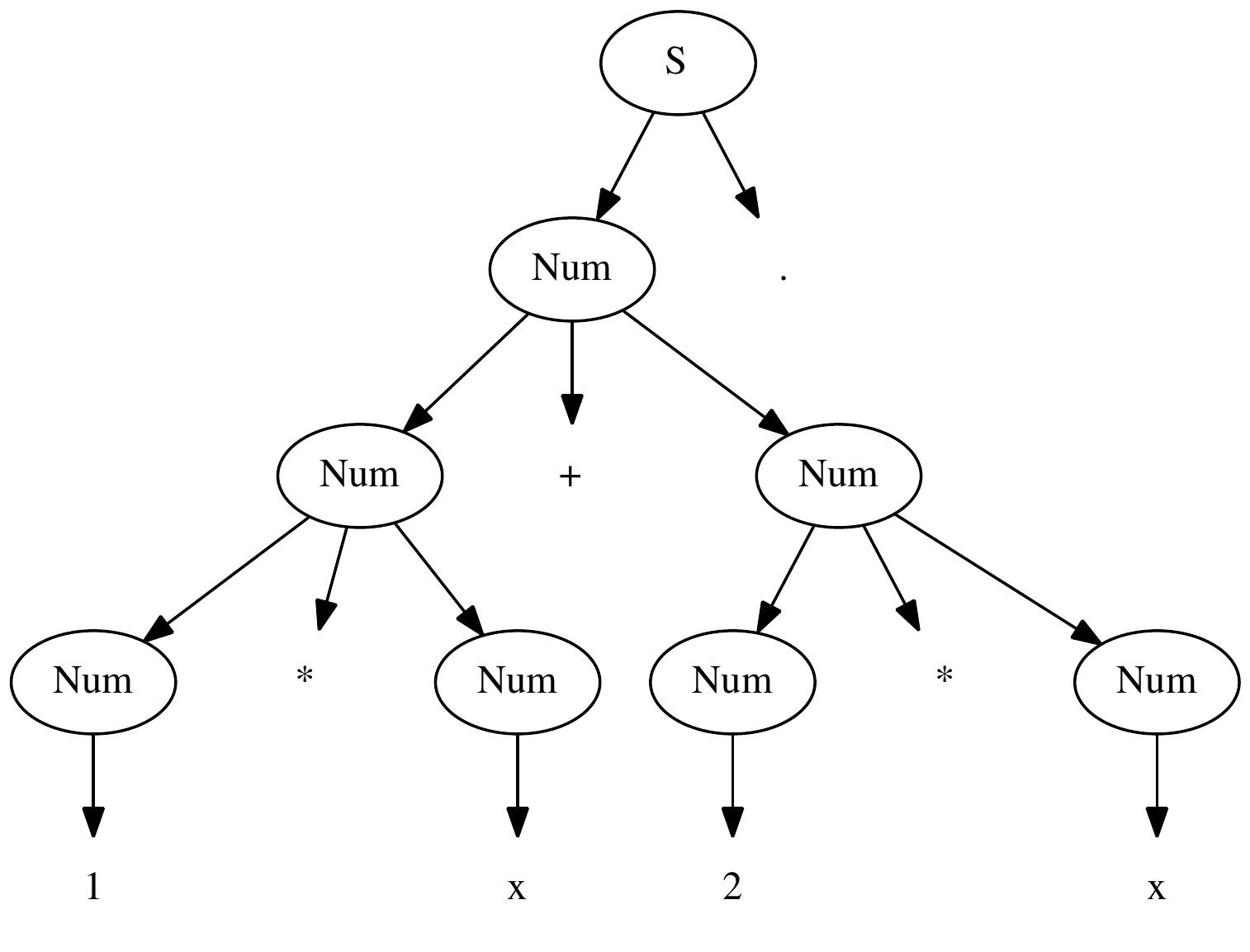}
\caption{The grammar tree $T_0(t)$.} 
\label{fig:example1}
 \end{figure}

\paragraph{Example:} 
Consider the following term $t$:
\begin{quote}
\begin{footnotesize}
\begin{verbatim}
1 * x + 2 * x.
\end{verbatim}
\end{footnotesize}
\end{quote}
with the following simplified parse tree $T_0(t)$ (see also Fig.~\ref{fig:example1}).
\begin{scriptsize}
\begin{verbatim}
(S (Num (Num (Num 1) * (Num x)) + (Num (Num 2) * (Num x))) .)
\end{verbatim}
\end{scriptsize}
When used as the training data (treebank), the grammar tree $T_0(t)$ results in the following set of CFG rules $G(T_0(t))$:
\begin{quote}
\begin{scriptsize}
\begin{verbatim}
S -> Num .                         Num -> 1 
Num ->  Num + Num                  Num -> 2 
Num -> Num * Num                   Num -> x 
\end{verbatim}
\end{scriptsize}
\end{quote}

This grammar allows exactly the following five parse trees $T_4(t), ..., T_0(t)$ when used on the original (non-bracketed) term $t$:

\begin{scriptsize}
\begin{verbatim}
(S (Num (Num 1) * (Num (Num (Num x) + (Num 2)) * (Num x))) .)
(S (Num (Num 1) * (Num (Num x) + (Num (Num 2) * (Num x)))) .)
(S (Num (Num (Num 1) * (Num (Num x) + (Num 2))) * (Num x)) .)
(S (Num (Num (Num (Num 1) * (Num x)) + (Num 2)) * (Num x)) .)
(S (Num (Num (Num 1) * (Num x)) + (Num (Num 2) * (Num x))) .)
\end{verbatim}
\end{scriptsize}

Here only the last tree corresponds to the original training
tree $T_0(t)$. No matter what probabilities $p(Rule_i)$ are assigned to the
grammar rules $G(T_0(t))$, it is not possible to make the priority of \texttt{+} smaller than
the priority of \texttt{*}. A context-free grammar forgets the context
and cannot remember and apply complex mechanisms such as
priorities. The probability of all parse trees is thus in this case always
the same, and equal to:
\begin{scriptsize}
\begin{align*}
p(T_4(t)) = ... = p(T_0(t)) =
p(\texttt{S -> Num .}) \times p(\texttt{Num -> Num + Num})\\ \times p(\texttt{Num -> Num * Num}) \times p(\texttt{Num -> Num * Num})\\  \times 
 p(\texttt{Num -> 1}) \times p(\texttt{Num -> 2}) \times p(\texttt{Num -> x}) \times p(\texttt{Num -> x})
\end{align*}
\end{scriptsize}

While the example's correct parse does not strictly imply the priorities of \texttt{+} and \texttt{*} as we know
them, it is clear that we would like the grammar to prefer parse trees that are in some
sense \emph{more similar} to the training data. One method that is
frequently used for dealing with similar problems in the NLP domain is
\emph{grammar
  lexicalization}~\cite{DBLP:conf/acl/Collins97}.
There an additional terminal can be appended to nonterminals and
propagated from the subtrees, thus creating many more possible (more
precise) nonterminals.  This approach however does not solve the
particular problem with operator priorities. We also believe that
considering probabilities of larger subtrees in the data as we propose
below is conceptually cleaner than lexicalization.

\section{Using Probabilities of Deeper Subtrees}\label{sec:subtrees}

Our solution is motivated by an
analogy with the n-gram statistical
machine-translation models, and also with the large-theory premise
selection systems. In such systems, characterizing formulas by all deeper
subterms and subformulas is feasible and typically considerably improves the
performance of the algorithms~\cite{ckjujv-ijcai15}. Considering
subtrees of greater depth for updating the parsing probabilities may initially seem
computationally involved. Below we however show that by using efficient
ATP-style indexing datastructures such as discrimination trees, this
approach becomes feasible, solving in a reasonably clean way some
of the inherent problems of the context-free grammars mentioned above.

In more detail, our approach is as follows. We extract not just
subtrees of depth 2 from the treebank (as is done by the standard PCFG), but all
subtrees up to a certain depth. 
Other approaches -- such as frequency-based rather than depth-based --
are possible. During the (modified) CYK chart parsing, the probabilities
of the parsed subtrees are adjusted by taking into account the statistics of such deeper subtrees
extracted from the treebank. The extracted subtrees are technically treated as new ``grammar rules'' of the form:
\begin{small}
\begin{quote}
\it{root of the subtree} \tt{->} \it{list of the children of the subtree}
\end{quote}
\end{small}

Formally, for a treebank (set of trees) $\mathbb T$, we thus define
$G^n(\mathbb T)$ to be the grammar rules of depth $n$ extracted from
$\mathbb T$. The standard context-free grammar $G(\mathbb T)$ then
becomes $G^2(\mathbb T)$, and we denote by $G^{n,m}(\mathbb T)$ where $n \leq m$ the
union\footnote{In general, a grammar could pick only some subtree depths instead of their contiguous intervals, but we do not use such grammars now.} $G^n(\mathbb T) \cup ... \cup G^m(\mathbb T)$ .
The probabilities of these deeper grammar rules are again learned from the treebank.
Our current solution treats the nonterminals on the left-hand sides as disjoint from the
old (standard CFG) nonterminals when counting the probabilities (this can be made more complicated in the future).
The
right-hand sides of such new grammar rules thus contain larger subtrees, allowing to 
compute the parsing probabilities using more context/structural information than in the standard context-free case.

For the example term $t$ from Section~\ref{sec:example} this works as follows. 
After the extraction of all subtrees of depth 2 and 3
and the appropriate adjustment of their probabilities, we get
a new extended set of probabilistic grammar rules $G^{2,3}(T_0(t)) \supset G(T_0(t))$. This grammar could again parse all the five
different parse trees $T_4(t), ... , T_0(t)$ as in Section~\ref{sec:example}, but now the 
probabilities $p(T_4(t)), ... , p(T_0(t))$ would in general differ, and an implementation would be able to choose the training tree $T_0(t)$ as the most
probable one. In the particular implementation that we use (see Section~\ref{sec:impl}) its probability is:
\begin{scriptsize}
\begin{align*}
p(T_0(t)) = p(\texttt{Num -> (Num 1)}) \times p(\texttt{Num -> (Num x)}) \times\\
p(\texttt{Num -> (Num 2)}) \times p(\texttt{Num -> (Num x)}) \times\\
p(\texttt{Num -> (Num Num * Num) + (Num Num * Num)}) \times\\
p(\texttt{S -> Num .})
\end{align*}
\end{scriptsize}
Here the second line from the bottom stands for the probability of a subtree of depth 3.
For the case of the one-element treebank ${T_0(t)}$, $p(T_0(t))$ would indeed be the highest probability. 
On the other hand, the probability of some of the other parses (e.g., $T_4(t)$ and $T_3(t)$ above) would remain unmodified, because in
such parses there are no subtrees of depth 3 from the training tree $T_0(t)$.

\subsection{Efficient Implementation of Deeper Subtrees}\label{sec:impl}
Discrimination trees~\cite{DBLP:books/el/RobinsonV01}, as first
implemented by Greenbaum~\cite{Greenbaum86}, index terms in a trie,
which keeps single path-strings at each of the indexed terms.  A
discrimination tree can be constructed efficiently, by inserting 
terms in the traversal preorder. Since discrimination trees are based
on path indexing, retrieval of matching subtrees during the
parsing is straightforward.

We use a discrimination tree $D$ to store all the subtrees $G^{n,m}(\mathbb T)$ from the
treebank $\mathbb T$ and to efficiently retrieve them together with their probabilities
during the chart parsing. The efficiency of the  implementation is important,
as we need to index about half a million subtrees in  $D$ for the experiments over Flyspeck.
On the other hand, such numbers
have become quite common in large-theory reasoning recently and do not
pose a significant problem. For memory efficiency we use OCaml maps
(implemented as AVL trees) in the internal nodes of $D$. The
lookup time thus grows logarithmically with the number of trees in
$D$, which is the main reason why we so far only consider trees of depth 3.

When a
particular cell in the CYK parsing chart is finished (i.e., all its
possible parses are known), the subtree-based probability update is
initiated.  The algorithm thus consists of two phases: (i) the
standard collecting of all possible parses of a particular cell, using
the context-free rules $G^2(\mathbb T)$ only, and (ii) the computation of
probabilities, which involves also the deeper (contextual) subtrees $G^{3,m}(\mathbb T)$.
 
In the second phase, every parse $P$ of the particular cell is
inspected, trying to find its top-level subtrees of depths $3, ..., m$ in the
discrimination tree $D$. If a matching tree $T$ is found in $D$, the
probability of $P$ is recomputed, using the probability of $T$. There
are various ways how to combine the old context-free and the new
contextual probabilities. The current method we use is to take the
maximum of the probabilities, keeping them as competing methods. As mentioned
above, the nonterminals in the new subtree-based rules are kept
disjoint from the old context-free rules when computing the grammar
rule probabilities. The usual effect is that a frequent deeper subtree
that matches the parse $P$ gives it more probability, because such a
``deeper context parse'' replaces the corresponding two shallow (old
context-free) rules, whose probabilities would have to be multiplied.


Our speed measurement with depth 3  has shown that the new implementation is
(surprisingly) 
faster. In particular, when training
on all 21695 Flypeck trees and testing on 11911 of them with the limit of 10 best parses, the new
version is 23\% faster than the old one (10342.75 s vs. 13406.97 s
total time). In this measurement the new version also failed to produce at least a single
parse less often than the old version (631 vs 818). 
This likely means that the deeper subtrees help to promote the correct parse,
which in the context-free version is considered at some point too improbable to
make it into the top 10 parses and consequently discarded.

\vspace{-1mm}
\section{Experimental Evaluation}\label{sec:evaluation}

\subsection{Machine Learning Evaluation}

The main evaluation is done in the same cross-validation scenario as in~\cite{KaliszykUV15}. 
We create the ambiguous sentences (Sec.~\ref{setting}) and the
disambiguated grammar trees from all 21695 Flyspeck
theorems,\footnote{About 1\% of the longest Flyspeck formulas were
  removed from the evaluation to keep the parsing times manageable.}
permute them randomly and split into 100 equally sized chunks of about
217 trees and their corresponding sentences.  The grammar trees serve for
training and the ambiguous sentences for evaluation.  For each testing chunk
$C_i$ ($i \in {1 .. 100}$) of 217 sentences we train the probabilistic
grammar $P_i$ on the union of the remaining 99 chunks of grammar trees
(altogether about 21478 trees).  Then we try to get the best 20 parse
trees for all the 217 sentences in $C_i$ using the grammar $P_i$.  This
is done for the simple context-free version (depth 2) of the algorithm
(Section~\ref{sec:align}), as well as for the versions using deeper
subtrees (Section~\ref{sec:subtrees}). 
The numbers of correctly parsed formulas
and their average ranks across the several 100-fold cross-validations
are shown in Table~\ref{Table2}.



\begin{table}[h!]
\begin{small}

\centering
\begin{tabular}{cccc}
\toprule
 depth  & correct parse found (\%) & avg. rank of correct parse \\\midrule
                           2                        & 8998 (41.5)      & 3.81                      \\
                           3                        & 11003 (50.7)     & 2.66                      \\
                           4                        & 13875 (64.0)     & 2.50                      \\
                           5                        & 14614 (67.4)     & 2.34                      \\
                           6                        & 14745 (68.0)     & 2.13                      \\
                           7                        & 14379 (66.2)     & 2.17                      \\\bottomrule
\end{tabular}
\caption{\label{Table2}\small{Numbers of correctly parsed Flyspeck theorems within first 20 parses and their average ranks for subtree depths 2 to 7 of the parsing algorithm (100-fold cross-validation)}.}
\end{small}
\end{table}

It is clear that the introduction of deeper subtrees into the CYK
algorithm has produced a significant improvement of the parsing
precision. The number of correctly parsed formulas appearing among the
top 20 parses has increased by 22\% between the context-free (depth 2)
version and the subtree-based version when using subtrees of depth 3,
and it grows by 64\% when using subtrees of depth 6.


The comparison of the average ranks is in general only a heuristic
indicator, because the number of correct parses found differ so
significantly between the methods.\footnote{If the context-free
  version parsed only a few terms, but with the best rank, its
  average rank would be 1, but the method would still be much worse in
  terms of the overall number of correctly parsed terms.} However,
since the number of parses is higher in the better-ranking methods,
this improvement is also relevant. The average rank of the best
subtree-based method (depth 6) is only about 56\% of the context-free
method.  The results of the best method say that for 68\% of the
theorems the correct parse of an ambiguous statement is among the best
20 parses, and its average rank among them is 2.13.



\vspace{-1mm}
\subsection{ATP Evaluation}

In the ATP evaluation we measure how many of the correctly parsed
formulas the HOL(y)Hammer system can prove, and thus help to confirm
their validity.  While the machine-learning evaluation is for simplicity
done by randomization, regardless of the chronological order of the
Flyspeck theorems, in the ATP evaluation we only allow facts that were
already proved in Flyspeck before the currently parsed
formula. Otherwise the theorem-proving task becomes too easy, because
the premise-selection algorithm will 
likely select the theorem
itself as the most relevant premise.
Since this involves large amount of computation, we only compare the
best new subtree-based method (depth 6) from Table~\ref{Table2}
(\emph{subtree-6}) with the old context-free method 
(\emph{subtree-2}).

In the ATP evaluation, the number of the Flyspeck theorems is reduced
from 21695 to 17018. This is due to omitting definitions and
duplicities during the chronological processing and ATP problem
generation.  For actual theorem proving, we only use a single
(strongest) HOL(y)Hammer method: the distance-weighted $k$-nearest neighbor
($k$-NN)~\cite{DudaniS76} 
using the strongest combination of
features~\cite{ckjujv-ijcai15}, with IDF-based feature
weighting~\cite{EasyChair:74} 
and 128 premises, and running Vampire
4.0~\cite{Vampire}. Running the full portfolio of 14 AI/ATP HOL(y)Hammer
strategies for hundreds of thousands problems would be too
computationally expensive.

\begin{table}[h!]
\begin{small}
\centering
\begin{tabular}{ccc}
  \toprule
                                                     & subtree-2   (\%)                                   & subtree-6  (\%)   \\\midrule
at least one parse (limit 20)                                  & 14101 (82.9)         & 16049      (94.3)     \\
at least one correct parse                           & 5744 (33.8)          & 10735    (63.1)      \\
at least one OLT parse                               & 808  (4.7)           & 1584      (9.3)       \\
at least one parse proved                            & 5682  (33.3)           & 7538  (44.3)                         \\
correct parse proved                                 & 1762    (10.4)           & 2616  (15.4)                        \\
at least one OLT parse proved                        & 525  (3.1)           & 814     (4.8)                      \\
the first parse proved is correct                    & 1168   (6.7)                                   & 2064  (12.1)                        \\
the first parse proved is OLT                        & 332   (2.0)                                      & 713  (4.2)                       \\\bottomrule
\end{tabular}
\caption{\label{Table2ATP}\small{Statistics of the ATP evaluation for subtree-2 and subtree-6. The total number of theorems tried is 17018 and we require 20 best parses. OLT stands for other library theorem.}}

\end{small}
\end{table}

Table~\ref{Table2ATP} shows the results. In
this evaluation we also detect situations when an ambiguated Flyspeck
theorem $T_1$ is parsed as a different known Flyspeck theorem
$T_2$. We call the latter situation \emph{other library theorem (OLT)}.
The removal of definitions and duplicitites made the difference in the
top-20 correctly parsed sentences even higher, going from 33.8\% for
subtree-2 to 63.1\% in subtree-6. This is an 81\% improvement. 
 A correspondingly high increase
between subtree-2 and subtree-6 is also in the number of situations
when the first parse is correct (or OLT) and HOL(y)Hammer can prove it
using the previous Flyspeck facts. The much greater easiness of
proving an existing library theorem than proving a new theorem
explains the relatively high number of provable OLTs when compared to
their total number of occurences. Such OLT proofs are however very
easy to filter out when using HOL(y)Hammer as a semantic filter for
the informal-to-formal translation.

\vspace{-2mm}
\section{Conclusion and Future Work}\label{sec:conclusion}

In comparison to the first results of~\cite{KaliszykUV15}, we have
very significantly increased the success rate of the informal-to-formal translation task on the Flyspeck corpus.
The overall improvement in the number of
correct parses among the top 20 is 64\% and even higher when omitting duplicities and definitions (81\%). The average rank of the
correct parse has decreased by 44\%. 
We believe that the contextual approach to enhancing CYK we took is
rather natural (in particular more natural than lexicalization), the
discrimination tree indexing scales to this task, and the performance
increase is very impressive. 

Future work includes adding further semantic checks and better
probabilistic ranking subroutines directly into the parsing process.
The chart-parsing algorithm is easy to extend with such checks and
subroutines, and already the current semantic pruning of parse trees
that have incompatible variable types is extremely important. While
some semantic relations might eventually be learnable by methods such
as recurrent neural networks (RNNs), we believe that the current
approach allows more flexible experimenting and nontrivial integration
and feedback loops between advanced deductive and learning components. A
possible use of RNNs in such a setup is for better ranking of subtrees
and for global focusing of the parsing process. 

An example of a more sophisticated deductive algorithm that should be
easy to integrate is congruence closure over provably equal (or
equivalent) parsing subtrees. For example, \texttt{``a * b * c''} can be
understood with different bracketing, different types of the variables
and different interpretations of \texttt{*}. However, \texttt{*} is
almost always associative across all types and interpretations. Human
readers know this, and rather than considering differently bracketed
parses, they focus on the real problem, i.e., which types to assign to
the variables and how to interpret the operator in the current
context. To be able to emulate this ability, we would cache directly in
the chart parsing algorithm the results of large-theory ATP runs on
many previously encountered equalities, and use them for fast
congruence closure over the subtrees.

\begin{small}
\bibliography{ate11}

\begin{thebibliography}{}

\bibitem[\protect\citeauthoryear{Bancerek and Rudnicki}{2002}]{BancerekR02}
Bancerek, G., and Rudnicki, P.
\newblock 2002.
\newblock A {Compendium of Continuous Lattices} in {MIZAR}.
\newblock {\em J. Autom. Reasoning} 29(3-4):189--224.

\bibitem[\protect\citeauthoryear{Blanchette \bgroup et al\mbox.\egroup
  }{2016}]{hammers4qed}
Blanchette, J.~C.; Kaliszyk, C.; Paulson, L.~C.; and Urban, J.
\newblock 2016.
\newblock Hammering towards {QED}.
\newblock {\em J. Formalized Reasoning} 9(1):101--148.

\bibitem[\protect\citeauthoryear{Collins}{1997}]{DBLP:conf/acl/Collins97}
Collins, M.
\newblock 1997.
\newblock Three generative, lexicalised models for statistical parsing.
\newblock In Cohen, P.~R., and Wahlster, W., eds., {\em 35th Annual Meeting of
  the Association for Computational Linguistics and 8th Conference of the
  European Chapter of the Association for Computational Linguistics,
  Proceedings of the Conference, 7-12 July 1997, Universidad Nacional de
  Educaci{\'{o}}n a Distancia (UNED), Madrid, Spain.},  16--23.
\newblock Morgan Kaufmann Publishers / {ACL}.

\bibitem[\protect\citeauthoryear{coq}{}]{coq}
The {Coq Proof Assistant}.
\newblock \url{http://coq.inria.fr}.

\bibitem[\protect\citeauthoryear{Deerwester \bgroup et al\mbox.\egroup
  }{1990}]{DeerwesterDLFH90}
Deerwester, S.~C.; Dumais, S.~T.; Landauer, T.~K.; Furnas, G.~W.; and Harshman,
  R.~A.
\newblock 1990.
\newblock {Indexing by Latent Semantic Analysis}.
\newblock {\em JASIS} 41(6):391--407.

\bibitem[\protect\citeauthoryear{Dudani}{1976}]{DudaniS76}
Dudani, S.~A.
\newblock 1976.
\newblock The distance-weighted k-nearest-neighbor rule.
\newblock {\em Systems, Man and Cybernetics, IEEE Transactions on}
  SMC-6(4):325--327.

\bibitem[\protect\citeauthoryear{Garillot \bgroup et al\mbox.\egroup
  }{2009}]{GarillotGMR09}
Garillot, F.; Gonthier, G.; Mahboubi, A.; and Rideau, L.
\newblock 2009.
\newblock Packaging mathematical structures.
\newblock In Berghofer, S.; Nipkow, T.; Urban, C.; and Wenzel, M., eds., {\em
  Theorem Proving in Higher Order Logics, 22nd International Conference, TPHOLs
  2009, Munich, Germany, August 17-20, 2009. Proceedings}, volume 5674 of {\em
  Lecture Notes in Computer Science},  327--342.
\newblock Springer.

\bibitem[\protect\citeauthoryear{Gonthier and Tassi}{2012}]{GonthierT12}
Gonthier, G., and Tassi, E.
\newblock 2012.
\newblock A language of patterns for subterm selection.
\newblock In Beringer, L., and Felty, A.~P., eds., {\em Interactive Theorem
  Proving - Third International Conference, {ITP} 2012, Princeton, NJ, USA,
  August 13-15, 2012. Proceedings}, volume 7406 of {\em Lecture Notes in
  Computer Science},  361--376.
\newblock Springer.

\bibitem[\protect\citeauthoryear{Gonthier \bgroup et al\mbox.\egroup
  }{2013}]{DBLP:conf/itp/GonthierAABCGRMOBPRSTT13}
Gonthier, G.; Asperti, A.; Avigad, J.; Bertot, Y.; Cohen, C.; Garillot, F.;
  Roux, S.~L.; Mahboubi, A.; O'Connor, R.; Biha, S.~O.; Pasca, I.; Rideau, L.;
  Solovyev, A.; Tassi, E.; and Th{\'e}ry, L.
\newblock 2013.
\newblock A machine-checked proof of the {Odd Order Theorem}.
\newblock In Blazy, S.; Paulin-Mohring, C.; and Pichardie, D., eds., {\em ITP},
  volume 7998 of {\em LNCS},  163--179.
\newblock Springer.

\bibitem[\protect\citeauthoryear{Grabowski, Korni{\l}owicz, and
  Naumowicz}{2010}]{mizar-in-a-nutshell}
Grabowski, A.; Korni{\l}owicz, A.; and Naumowicz, A.
\newblock 2010.
\newblock {M}izar in a nutshell.
\newblock {\em J. Formalized Reasoning} 3(2):153--245.

\bibitem[\protect\citeauthoryear{Greenbaum}{1986}]{Greenbaum86}
Greenbaum, S.
\newblock 1986.
\newblock {\em Input transformations and resolution implementation techniques
  for theorem-proving in first-order logic}.
\newblock Ph.D. Dissertation, University of Illinois at Urbana-Champaign.

\bibitem[\protect\citeauthoryear{Haftmann and Wenzel}{2006}]{HaftmannW06}
Haftmann, F., and Wenzel, M.
\newblock 2006.
\newblock Constructive type classes in isabelle.
\newblock In Altenkirch, T., and McBride, C., eds., {\em Types for Proofs and
  Programs, International Workshop, {TYPES} 2006, Nottingham, UK, April 18-21,
  2006, Revised Selected Papers}, volume 4502 of {\em Lecture Notes in Computer
  Science},  160--174.
\newblock Springer.

\bibitem[\protect\citeauthoryear{Hales \bgroup et al\mbox.\egroup
  }{2015}]{HalesABDHHKMMNNNOPRSTTTUVZ15}
Hales, T.~C.; Adams, M.; Bauer, G.; Dang, D.~T.; Harrison, J.; Hoang, T.~L.;
  Kaliszyk, C.; Magron, V.; McLaughlin, S.; Nguyen, T.~T.; Nguyen, T.~Q.;
  Nipkow, T.; Obua, S.; Pleso, J.; Rute, J.; Solovyev, A.; Ta, A. H.~T.; Tran,
  T.~N.; Trieu, D.~T.; Urban, J.; Vu, K.~K.; and Zumkeller, R.
\newblock 2015.
\newblock A formal proof of the {K}epler conjecture.
\newblock {\em CoRR} abs/1501.02155.

\bibitem[\protect\citeauthoryear{Hales}{2012}]{hales-dense}
Hales, T.
\newblock 2012.
\newblock {\em Dense Sphere Packings: A Blueprint for Formal Proofs}, volume
  400 of {\em London Mathematical Society Lecture Note Series}.
\newblock Cambridge University Press.

\bibitem[\protect\citeauthoryear{Harrison, Urban, and
  Wiedijk}{2014}]{HarrisonUW14}
Harrison, J.; Urban, J.; and Wiedijk, F.
\newblock 2014.
\newblock History of interactive theorem proving.
\newblock In Siekmann, J.~H., ed., {\em Computational Logic}, volume~9 of {\em
  Handbook of the History of Logic}. Elsevier.
\newblock  135--214.

\bibitem[\protect\citeauthoryear{Harrison}{1996}]{Harrison96}
Harrison, J.
\newblock 1996.
\newblock {HOL Light}: A tutorial introduction.
\newblock In Srivas, M.~K., and Camilleri, A.~J., eds., {\em FMCAD}, volume
  1166 of {\em LNCS},  265--269.
\newblock Springer.

\bibitem[\protect\citeauthoryear{Kaliszyk and Urban}{2013}]{EasyChair:74}
Kaliszyk, C., and Urban, J.
\newblock 2013.
\newblock Stronger automation for {F}lyspeck by feature weighting and strategy
  evolution.
\newblock In Blanchette, J.~C., and Urban, J., eds., {\em PxTP 2013}, volume~14
  of {\em EPiC Series},  87--95.
\newblock EasyChair.

\bibitem[\protect\citeauthoryear{Kaliszyk and Urban}{2014}]{holyhammer}
Kaliszyk, C., and Urban, J.
\newblock 2014.
\newblock Learning-assisted automated reasoning with {F}lyspeck.
\newblock {\em J. Autom. Reasoning} 53(2):173--213.

\bibitem[\protect\citeauthoryear{Kaliszyk, Urban, and
  Vyskocil}{2015a}]{ckjujv-ijcai15}
Kaliszyk, C.; Urban, J.; and Vyskocil, J.
\newblock 2015a.
\newblock Efficient semantic features for automated reasoning over large
  theories.
\newblock In Yang, Q., and Wooldridge, M., eds., {\em IJCAI'15},  3084--3090.
\newblock {AAAI} Press.

\bibitem[\protect\citeauthoryear{Kaliszyk, Urban, and
  Vyskocil}{2015b}]{KaliszykUV15}
Kaliszyk, C.; Urban, J.; and Vyskocil, J.
\newblock 2015b.
\newblock Learning to parse on aligned corpora (rough diamond).
\newblock In Urban, C., and Zhang, X., eds., {\em Interactive Theorem Proving -
  6th International Conference, {ITP} 2015, Nanjing, China, August 24-27, 2015,
  Proceedings}, volume 9236 of {\em Lecture Notes in Computer Science},
  227--233.
\newblock Springer.

\bibitem[\protect\citeauthoryear{Klein \bgroup et al\mbox.\egroup
  }{2010}]{KleinAEHCDEEKNSTW10}
Klein, G.; Andronick, J.; Elphinstone, K.; Heiser, G.; Cock, D.; Derrin, P.;
  Elkaduwe, D.; Engelhardt, K.; Kolanski, R.; Norrish, M.; Sewell, T.; Tuch,
  H.; and Winwood, S.
\newblock 2010.
\newblock {seL4}: formal verification of an operating-system kernel.
\newblock {\em Commun. ACM} 53(6):107--115.

\bibitem[\protect\citeauthoryear{Kov{\'a}cs and Voronkov}{2013}]{Vampire}
Kov{\'a}cs, L., and Voronkov, A.
\newblock 2013.
\newblock First-order theorem proving and {V}ampire.
\newblock In Sharygina, N., and Veith, H., eds., {\em CAV}, volume 8044 of {\em
  LNCS},  1--35.
\newblock Springer.

\bibitem[\protect\citeauthoryear{Lange and
  Lei{\ss}}{2009}]{DBLP:journals/didactica/LangeL09}
Lange, M., and Lei{\ss}, H.
\newblock 2009.
\newblock To {CNF} or not to {CNF}? an efficient yet presentable version of the
  {CYK} algorithm.
\newblock {\em Informatica Didactica} 8.

\bibitem[\protect\citeauthoryear{Leroy}{2009}]{Leroy09}
Leroy, X.
\newblock 2009.
\newblock Formal verification of a realistic compiler.
\newblock {\em Commun. {ACM}} 52(7):107--115.

\bibitem[\protect\citeauthoryear{Robinson and
  Voronkov}{2001}]{DBLP:books/el/RobinsonV01}
Robinson, J.~A., and Voronkov, A., eds.
\newblock 2001.
\newblock {\em Handbook of Automated Reasoning (in 2 volumes)}.
\newblock Elsevier and {MIT} Press.

\bibitem[\protect\citeauthoryear{Rudnicki, Schwarzweller, and
  Trybulec}{2001}]{RudnickiST01}
Rudnicki, P.; Schwarzweller, C.; and Trybulec, A.
\newblock 2001.
\newblock Commutative algebra in the {M}izar system.
\newblock {\em J. Symb. Comput.} 32(1/2):143--169.

\bibitem[\protect\citeauthoryear{Tankink \bgroup et al\mbox.\egroup
  }{2013}]{FlyspeckWiki}
Tankink, C.; Kaliszyk, C.; Urban, J.; and Geuvers, H.
\newblock 2013.
\newblock Formal mathematics on display: A wiki for {Flyspeck}.
\newblock In Carette, J.; Aspinall, D.; Lange, C.; Sojka, P.; and Windsteiger,
  W., eds., {\em MKM/Calculemus/DML}, volume 7961 of {\em LNCS},  152--167.
\newblock Springer.

\bibitem[\protect\citeauthoryear{Wenzel, Paulson, and
  Nipkow}{2008}]{WenzelPN08}
Wenzel, M.; Paulson, L.~C.; and Nipkow, T.
\newblock 2008.
\newblock The {I}sabelle framework.
\newblock In Mohamed, O.~A.; Mu{\~n}oz, C.~A.; and Tahar, S., eds., {\em
  TPHOLs}, volume 5170 of {\em LNCS},  33--38.
\newblock Springer.

\bibitem[\protect\citeauthoryear{Younger}{1967}]{Younger67}
Younger, D.~H.
\newblock 1967.
\newblock Recognition and parsing of context-free languages in time n{\^{}}3.
\newblock {\em Information and Control} 10(2):189--208.

\bibitem[\protect\citeauthoryear{Zinn}{2004}]{Zinn2004}
Zinn, C.
\newblock 2004.
\newblock {\em Understanding informal mathematical discourse}.
\newblock Ph.D. Dissertation, University of Erlangen-Nuremberg.

\end{thebibliography}
\bibliographystyle{aaai}
\end{small}
\end{document}